\documentclass[11pt]{article}
\usepackage[T1]{fontenc}
\usepackage[margin=1in]{geometry}
\usepackage{amsmath,amssymb,amsfonts}
\usepackage{graphicx}
\usepackage{booktabs}
\usepackage{xcolor}
\usepackage[round,authoryear]{natbib}
\usepackage[colorlinks=true,linkcolor=blue!60!black,citecolor=blue!60!black,urlcolor=blue!60!black]{hyperref}
\usepackage{caption}
\usepackage{microtype}
\hypersetup{
  pdftitle={Portable Semantics, Private Dialects: Reuse and Negative Transfer in Latent Communication Between Language-Model Cells},
  pdfauthor={Narcis Marincat},
  pdfsubject={Cross-model interoperability and transfer value of emergent latent communication interfaces}
}

\newcommand{\ckpt}[1]{\texttt{#1}}

\title{Portable Semantics, Private Dialects:\\ Reuse and Negative Transfer in Latent Communication Between Language-Model Cells}
\author{Narcis Marincat\thanks{Independent researcher. Correspondence: \texttt{narcis.marincat.20@ucl.ac.uk}.}}
\date{August 2026}

\begin{document}
\maketitle

\begin{abstract}
In shared-genome language-model societies, restricted evidence visibility favors reusable, value-indexed latent packet interfaces, whereas the sole high-performing globally visible model in the parent study learned an episode-entangled code \citep{marincat2026populus}. This companion study asks three second-order questions about such emergent interfaces: whether independently trained societies share one packet language, where strict zero-shot transfer fails, and whether inherited interface state helps or harms later learning. First, a leakage-controlled causal interoperability audit over all 30 ordered off-diagonal pairs of six independently trained restricted societies---with sealed held-out values, programs, phrasings, and packet positions, and a preregistered raw/orthogonal/linear/nonlinear alignment ladder---shows that the six semantically similar interfaces do not form one raw language: the only raw interoperability observed lies within same-initialization strata. One same-initialization pair is exactly interoperable in both directions, a second shows asymmetric partial compatibility, and all 26 cross-initialization directions fail every frozen alignment rung. Second, within the tested decomposition and the single sealed Family-D source formulation, a source-span control localizes strict zero-shot failure in the restricted societies to interpretation and execution of the new operator instructions, rather than to source-state parsing, packet carriage, or output decoding. Third, in a matched adaptation factorial, the globally trained communication interface acts as a severe negative-transfer prior: reinitializing only the packet reader, writer, and mouth---while retaining the same globally trained low-rank adapter, target stream, and budget---raises final depth-three accuracy from 0.169 to 0.857, and the frozen inherited interface remains at chance. Fourth, across two restricted checkpoints and two independently frozen target streams each, inherited interfaces never exceeded fresh-interface controls by the preregistered 0.10 margin. Together the results indicate that portable semantics need not share raw coordinates, and that episode-entangled communication can become a severe negative-transfer prior. All primary conclusions are bounded to a near-transfer 17-state setting; the negative-transfer factorial concerns one globally visible parent-cohort checkpoint, while Appendix~A adds a separate post hoc tagged-global twin case.
\end{abstract}

\section{Introduction}\label{sec:intro}

The parent study \citep{marincat2026populus} trained four-cell societies of language-model cells that share one frozen pretrained genome and one low-rank adapter and communicate only through learned continuous packets. Its central causal result was that restricting each cell's evidence visibility substantially increases the probability of learning a generalizing communication relay, and that the relays learned under restriction repeatedly expose an approximately \emph{value-indexed} packet interface: natural packets are interchangeable within a checkpoint whenever they carry the same running intermediate value. The sole high-performing globally visible model also depended on its channel, but its packets were not interchangeable across episodes---an \emph{episode-entangled} code. A subsequent portability scout (Appendix~F of the parent paper) showed that each restricted interface transports its value code perfectly into a new instruction family, while strict zero-shot transfer of the whole frozen society fails.

Those results are all within-checkpoint or within-lineage. They leave open three second-order questions that matter for any programme that hopes to \emph{reuse} emergent latent interfaces---for example, by pretraining communication before task training, or by composing separately trained agents:

\begin{enumerate}
\item \textbf{Interoperability.} Do independently trained societies converge on one shared packet language, or on private dialects? If dialects, are they related by simple transformations?
\item \textbf{Failure localization.} When a frozen society fails zero-shot on a new instruction family, which stage fails---parsing the new source statement, carrying the value, executing the new operators, or decoding the output?
\item \textbf{Transfer value.} Does inheriting a trained interface accelerate learning of a new family, do nothing, or actively harm it?
\end{enumerate}

We answer these with a preregistered four-part campaign (P1--P4) run on the parent study's released checkpoints under sealed manifests, frozen decision rules, and bit-exact machine admission. P1 is a causal cross-model transplant audit over all 30 ordered off-diagonal pairs of the six audited restricted societies. P2 is a source-span control that isolates the parsing-and-publication stage. P3 is a matched component-reinitialization factorial on the globally visible checkpoint. P4 is a second-stream replication of the interface-inheritance comparison for restricted checkpoints.

\paragraph{Contributions.}
\begin{enumerate}
\item \textbf{A leakage-controlled causal interoperability audit for emergent latent communication.} Across every directed pair of six independently trained societies, we test raw and learned packet alignments on semantic values and packet positions excluded from fitting, using natural packet injection and derangement controls (\S\ref{sec:p1}).
\item \textbf{Initialization-stratified dialects.} The learned value interfaces are semantically similar but do not form one raw language. Raw interoperability is exact for one same-initialization pair and partial for another, whereas all cross-initialization directions fail every frozen alignment rung---consistent with initialization-anchored, training-perturbed coordinate dialects (\S\ref{sec:p1results}).
\item \textbf{Interface-level negative transfer.} In a matched adaptation factorial, the sole successful globally visible model in the parent cohort has an episode-entangled interface that acts as a severe negative-transfer prior: reinitializing only the reader, writer, and mouth while retaining the same LoRA raises target accuracy from 0.169 to 0.857 (\S\ref{sec:p3}). This is a checkpoint-specific mechanistic case study, not a population-level estimate for globally trained interfaces.
\end{enumerate}

Two additional results sharpen these contributions. A second independently frozen target stream for each restricted checkpoint replicates the absence of the preregistered large inherited-interface benefit while leaving the original failed criterion unchanged (\S\ref{sec:p4}), and P2 localizes the zero-shot failure within the target computation (\S\ref{sec:p2}).

\paragraph{Novelty, stated exactly.}
Prior work studies zero-shot coordination between independently trained agents, cross-model representation alignment, causal interchange interventions, and negative transfer from warm starts. To our knowledge, this is the first study to combine a sealed causal interoperability audit of an emergent neural communication interface with a matched component-reinitialization factorial that localizes negative transfer to the learned communication interface itself. We do not claim the first causal cross-model transplant, the first functional model alignment, the first orthogonal alignment for latent agent communication, or the first derangement-controlled communication audit; \S\ref{sec:related} details the precedents for each ingredient.

\section{Setting and prior results}\label{sec:setting}

\paragraph{Societies.} A society is four cells in a fixed relay sharing one frozen Qwen2.5-0.5B-Instruct genome \citep{qwen25} and one rank-8 LoRA \citep{hu2022lora}. Each cell reads its assigned evidence span and the incoming packet, and commits an outgoing packet of two model-width (896-dimensional) continuous vectors---the only inter-cell information pathway; the final cell decodes through a frozen LM head plus a learned mouth projection. The task is ordered natural-language function composition over $\mathbb{Z}_{17}$ (chance $=1/17\approx0.0588$); evaluation uses held-out operator programs in held-out phrasings. In the \emph{restricted} arm each cell may attend only to its own evidence span; in the \emph{global} arm all spans are readable. Checkpoints are named by initialization and data order: for example \ckpt{m204\_o954} is initialization 204 trained under order 954. The six audited restricted checkpoints are \ckpt{m200\_o950}, \ckpt{m201\_o951}, \ckpt{m203\_o903}, \ckpt{m203\_o953}, \ckpt{m204\_o904}, and \ckpt{m204\_o954}; the sole high-performing globally visible checkpoint (denoted G) shares initialization 204 and order 954.

\paragraph{Parent results used here.} All six audited restricted societies expose approximately value-indexed packet interfaces: within a checkpoint, natural same-value packet transplants preserve behavior at 0.94--1.00 across all tested interfaces, and counterfactual-value packets redirect outputs toward the mathematically predicted answer. The G comparator requires its channel but its same-value packets are not interchangeable across episodes \citep{marincat2026populus}.

\paragraph{The portability scout (parent Appendix F).} A new instruction family (Family D) defines a fresh set of affine transformations over the same $\mathbb{Z}_{17}$ carrier space, with zero exact primitive-map overlap with Family A, a disjoint operator vocabulary, and a new phrasing distribution. Strict zero-shot transfer fails at chance for all seven checkpoints, yet the failure is sharply structured: transplanting each of the six audited restricted societies' Family-A value packets into Family-D episodes preserves behavior at 1.000 (carrier portability), mouth decoding ports at 1.000, and an exhaustive grid shows no new-operator execution. The family is learnable from scratch (a fresh society reaches 0.9167 depth-three accuracy in 20{,}000 updates, passing 0.40 by 2{,}500), and an interface-inheritance probe in two purposively selected restricted checkpoints found that a frozen inherited interface supports Family-D learning, while inheriting interface state conferred no preregistered $\geq 0.10$ advantage over a fresh interface. P1--P4 turn these observations into sealed, controlled comparisons.

\paragraph{Campaign discipline.} Experiments P1--P4 were frozen in a written mini-plan before execution. P1 additionally ran under a launch manifest committed before packet harvest, fixing the value and programme splits, derangements, mapper specifications and seeds, checkpoint hashes, and the classification script; its sealed manifest recorded the fit-bank, test-bank, and map hashes before any held-out score was computed. P2--P4 used the checkpoint identities, target-stream seeds, arms, budgets, thresholds, and interpretation rules committed in the frozen mini-plan. Evaluation machines were admitted only after bit-exact retrace of a golden training segment and matching fresh-world fingerprints.

\section{Related work}\label{sec:related}

\paragraph{Causal interchange and cross-model alignment.}
Causal abstraction tests whether internal states realize high-level variables via interchange interventions \citep{geiger2021causal}. Model Alignment Search \citep{grant2025mas} learns invertible cross-network mappings that causally transfer variable values, including numerosity, between networks trained from different seeds---the sharpest precedent against any broad ``first causal cross-model transfer'' claim, which we accordingly do not make. Functional cross-model compatibility likewise predates this work through transformation and stitching layers \citep{lenc2014equivariance,bansal2021stitching} and relative representations designed to remove seed-dependent rotations \citep{moschella2022relative}; \citet{almudevar2026usable} connect functional and representational similarity and highlight mapper-capacity dependence, relevant to our alignment ladder. What distinguishes P1 is the conjunction: the transplanted states are messages learned expressly for communication; the high-level variable (the running $\mathbb{Z}_{17}$ value) is known exactly; interoperability is tested over all directed pairs of independently trained systems; the receiver is tested behaviorally by injecting the donor's natural packet into its actual downstream computation; alignment fitting is segregated from evaluation by held-out values, programs, phrasings, positions, and disjoint packet instances; maps are frozen non-adaptively before scores are opened; and correct injections are checked against fixed-point-free derangements and shuffled controls. The derangement is a validation feature, not a methodological first.

\paragraph{Latent communication between language models.}
Recent systems communicate through hidden states rather than text. A universal activation interface learns one adapter pair per model \citep{kim2026adapter}; StateBridge aligns sender states to a receiver's input space with a training-free closed-form orthogonal transformation \citep{peng2026statebridge}, preempting any claim that P1 is the first Procrustes-style alignment for latent agent communication. XBridge identifies an entity-grounding failure in continuous bridges \citep{yang2026xbridge}. Causal audits of latent channels use mismatched, deranged, zeroed, and moment-matched controls \citep{zhang2026latent,cheng2026kv}, and \citet{truong2026gauge} combine disjoint alignment and probe data with a leakage audit for donor-specific fingerprints, while explicitly limiting their claim to detectability rather than transplantability or causal persistence. \citet{zhang2026pythia} report that high geometric alignment need not yield causal cross-model transfer, and \citet{cardenas2026causal} find preregistered causal activation transfer succeeding in only one of three decoder pairs despite comparable representational alignment---both directly reinforcing our distinction between geometric alignment and functional packet interoperability. Hidden APIs \citep{ma2026hidden} identify reusable causal interfaces within monolithic models; we study interfaces learned expressly for inter-cell communication.

\paragraph{Conventions, initialization, and cross-play.}
Emergent-communication research established that independently trained partners can converge on incompatible arbitrary conventions: Other-Play targets exactly this failure \citep{hu2020otherplay}, quasi-equivalence discovery seeks protocols that coordinate across independent runs \citep{bullard2021qed}, population heterogeneity shapes convention stability \citep{rita2022population}, inter-seed cross-play is itself only one level of robustness \citep{wolski2026crossplay}, and translations can be fitted between independently emerged protocols \citep{levy2025translation}. Initialization and training-randomness effects are similarly well established: networks sharing an initialization can become stable to SGD noise within a linearly connected region \citep{frankle2019lmc}, weight initialization and data order both materially affect fine-tuning outcomes \citep{dodge2020finetuning}, and permutation symmetries can conceal shared basins \citep{ainsworth2022rebasin}. Two close functional precedents are cross-seed autoencoder decoding collapse \citep{okatan2025keys} and seed-identifiable output fingerprints \citep{suzuki2025fingerprints}. To our knowledge, prior work has not jointly fixed the interface initialization, varied data order, directly injected one trained model's emergent messages into another, compared against cross-initialization injections, and evaluated exact semantic behavior; that functional protocol-interoperability stratification by initialization is what P1 contributes.

\paragraph{Negative transfer and plasticity.}
Source training can make target learning worse than starting afresh \citep{wang2018negative}; warm-started networks can generalize worse despite similar training losses \citep{ash2019warm}; pretrained networks can lose plasticity \citep{berariu2021plasticity,dohare2023plasticity}; early experience creates critical-period and primacy effects \citep{achille2017critical,nikishin2022primacy}; and partial reinitialization has been used to localize which components carry transfer \citep{tamkin2020transferability}. The closest new neighbor to P3 is Cross-Model Memory Transfer via Target-Side Reader Adaptation \citep{li2026memory}, which shows that transferred-memory utility depends on reader compatibility and that target-side reader adaptation can recover cross-model utility. P3 remains distinct because the inherited interface is not merely incompatible: retaining it produces severe negative transfer relative to reinitializing only the interface under a matched target-learning protocol, and full adaptation of the inherited interface does not escape (\S\ref{sec:p3}).

\section{P1: a sealed cross-model transplant audit}\label{sec:p1}

\subsection{Design}\label{sec:p1design}

P1 asks whether the six restricted societies' value-indexed interfaces are mutually intelligible: can a natural packet harvested from donor society $X$, indexed by the running value it carries, causally drive recipient society $Y$'s downstream computation?

\paragraph{Sealed banks.} For each checkpoint we harvest natural committed packets from its own episodes, indexed by (running value, packet position $k\in\{0,1,2,3\}$), with at least five instances per class and episode-level segregation into a \emph{fit} bank and a \emph{test} bank stored in separate files; checkpoint identity is verified by SHA-256 against the launch manifest.

\paragraph{Held-out structure.} The 17 values are split 9 fit / 8 test; operator programmes are partitioned 30/30; fit and test banks use disjoint constant phrasing templates; Global alignment maps are fitted only at packet positions $k\in\{1,2\}$. Separately, position-specific orthogonal, affine-linear, and nonlinear maps are fitted at each $k\in\{0,1,2,3\}$ on the nine fit values and evaluated on held-out values at that same position; scores on fit values at $k\in\{0,3\}$ are descriptive and do not gate position-specific success. For the global maps, test cells fall into three strata: \emph{semantic} (held-out values at fitting positions), \emph{position} (fit values at held-out positions $k\in\{0,3\}$), and \emph{joint} (held-out values at held-out positions). A fixed-point-free derangement over the test values defines counterfactual injections, and preregistered shuffle maps (a stored 3-shift over each value set) define the binding negative control.

\paragraph{Alignment ladder.} Each ordered pair receives a global-ladder classification according to the least flexible global rung whose complete gate set passes: \textbf{raw} (identity), \textbf{orthogonal} (float64 Procrustes with slot-mean removal and restoration), \textbf{affine linear} (centered float64 dual ridge over a $\lambda$ grid), or \textbf{nonlinear} (width-64 MLP; pass requires 2 of 3 frozen seeds). Position-specific versions of the three fitted map classes are evaluated separately as a diagnostic rescue and do not enter the same ordered ladder. Mapped packets are RMS-renormalized to the recipient's committed-packet scale before injection. Passing requires every cell $\geq0.90$, recipient self-diagonal $\geq0.95$, and shuffle $\leq0.11$ or $\geq0.30$ below the paired same-value score. Injections replace the recipient's committed packet at the corresponding packet position in batched evaluation; $k{=}3$ is the terminal packet presented to the mouth rather than an inter-cell edge. The primary context is instruction Family A; Family D is scored secondarily without refitting. The classification script was written and hashed into the sealed manifest before any held-out score was computed.

\paragraph{Instrumentation notes.} In this pure-carrier layout the same-value and counterfactual conditions are the same indexed-packet-following test under different target values, so we do not count them as independent replications; the shuffle control carries the burden of showing that the recipient follows packet identity rather than its local source span. At $k=0$ the source packet is deterministic per value (effectively one donor instance per class), a caveat recorded in the sealed classification output.

\subsection{Results}\label{sec:p1results}

\paragraph{Diagonals and controls.} Every recipient's self-diagonal is 1.000 in every stratum and both contexts. In the raw-passing pair, every shuffle control is 0.000. Across all directed pairs, shuffle controls average 0.052 at the raw rung and 0.006 across the fitted-map variants---nonzero values arise only in small held-out cells of failing directions---and no passing cell's shuffle approaches its paired indexed-packet score.

\begin{table}[t]
\centering
\caption{P1 same-initialization directions (raw rung, both contexts). The initialization-204 pair passes every cell at ceiling in both directions; the initialization-203 pair fails the preregistered every-cell $\geq0.90$ floor asymmetrically. Cells report same-value transfer accuracy; shuffle controls for the 204 pair are 0.000 in every cell.}
\label{tab:p1same}
\begin{tabular}{llccc}
\toprule
Direction & Verdict & Min cell & Weakest cell & Shuffle (max) \\
\midrule
\ckpt{m204\_o904} $\to$ \ckpt{m204\_o954} & raw pass & 1.000 & --- & 0.000 \\
\ckpt{m204\_o954} $\to$ \ckpt{m204\_o904} & raw pass & 1.000 & --- & 0.000 \\
\ckpt{m203\_o903} $\to$ \ckpt{m203\_o953} & fail & 0.875 & joint $k{=}0$ & 0.000 \\
\ckpt{m203\_o953} $\to$ \ckpt{m203\_o903} & fail & 0.750 & semantic $k{=}1$, joint $k{=}0$ & 0.000 \\
\bottomrule
\end{tabular}
\end{table}

\paragraph{One exact raw component.} The two initialization-204 societies are exactly raw-interoperable: every test cell in every stratum, both directions, and both instruction families scores 1.000 with shuffle 0.000 (Table~\ref{tab:p1same}). Trained under different data orders, these two societies nevertheless preserved a directly interchangeable packet convention---a clean existence result: a shared initialization can preserve a directly interoperable packet convention across different training orders.

\paragraph{Asymmetric partial compatibility.} The initialization-203 pair exhibits asymmetric partial raw interoperability: one direction narrowly misses the strict cellwise floor (minimum cell 0.875), whereas the reverse direction contains a more substantial localized failure (minimum cell 0.750) despite a high aggregate mean. Under the frozen rules this pair is not promoted to a pass, and we report it as partial compatibility rather than a near-miss.

\begin{table}[t]
\centering
\caption{P1 cross-initialization directions (all 26), held-out test cells pooled over both contexts. No rung passes any direction. Position-stratum cells use fitted values at held-out positions; semantic cells require extrapolation to held-out values.}
\label{tab:p1cross}
\begin{tabular}{lccc}
\toprule
Rung & Semantic cells (range / mean) & Position cells (range / mean) & Directions passing \\
\midrule
Raw & 0.000--0.250 / 0.041 & \multicolumn{1}{c}{---} & 0 / 26 \\
Orthogonal & 0.025--0.700 / 0.356 & 0.889--1.000 / 0.998 & 0 / 26 \\
Affine linear & 0.000--0.425 / 0.199 & \multicolumn{1}{c}{---} & 0 / 26 \\
Nonlinear (MLP-64) & 0.000--0.325 / 0.103 & \multicolumn{1}{c}{---} & 0 / 26 \\
\bottomrule
\end{tabular}
\end{table}

\paragraph{No cross-initialization alignment.} All 26 cross-initialization directions fail the raw and every fitted-alignment gate (Table~\ref{tab:p1cross}). The orthogonal maps often transfer fitted value classes across packet positions---position-stratum cells reach 0.889--1.000---indicating stage-general geometric structure, but they fail semantic extrapolation to held-out values. The affine and nonlinear rungs do no better on the semantic stratum, with the width-64 MLP degrading toward memorization of the fit classes. No tested global or position-specific low-complexity map makes cross-initialization packets functionally interchangeable on the sealed value classes. We emphasize what this does and does not establish: no tested raw, orthogonal, linear, or width-64 nonlinear alignment made cross-initialization packets functionally interchangeable on the sealed semantic classes; the codes are not thereby proven to be private ciphers, and cross-initialization alignment is not proven impossible.

\paragraph{Summary.} The six societies do not share one raw packet language. Raw compatibility is strongly initialization-stratified: one same-initialization pair is perfectly interoperable, another is partially interoperable, and all 26 cross-initialization directions fail every frozen alignment gate. Among the two initializations represented under two training orders, raw interoperability appeared only within initialization strata---exact for initialization 204 and substantial but subthreshold for initialization 203. This pattern is consistent with initialization anchoring the packet coordinate system more strongly than data order, but it does not identify initialization as the sole causal determinant of the learned code: with only two initializations under two orders each, and one pair only partially compatible, the certified reading is \emph{initialization-anchored, training-perturbed coordinate dialects}.

\section{P2: localizing the zero-shot failure}\label{sec:p2}

The scout established that strict zero-shot Family-D transfer fails at chance while the value carrier and mouth port perfectly. P2 closes the remaining gap: does the first cell correctly \emph{parse} the unfamiliar Family-D start-value statement and \emph{publish} the corresponding source packet?

For each checkpoint we present sealed source-only episodes---a Family-D start-value statement with no operators---and score the society's output over all 17 values, alongside a matched Family-A source control and a teacher-forced source-packet control. An additional source-span control showed that each of the six restricted-visibility checkpoints converted the sealed Family-D start-value statement into a functionally correct source packet for all 17 values (1.000 in the source-only carrier test); the Family-A source and teacher-forced packet controls were likewise 1.000. The globally visible comparator also solved the source-only episode at 1.000, although direct access to the source span makes that condition non-isolating for its first-cell packet. Together with ceiling packet and mouth portability and chance-level exhaustive operator execution, these results localize strict zero-shot failure in the restricted arm, within the tested decomposition, to interpretation and execution of the new operator instructions---not to source-state parsing, packet carriage, or final value-to-label decoding. The claim is bounded to the single sealed Family-D start-value formulation; it is not a paraphrase-generalization result.

\section{P3: a matched interface-adaptation factorial}\label{sec:p3}

\subsection{Design}

P3 asks whether the globally trained checkpoint's episode-entangled interface can be reused when its genome must learn Family D---and, if not, whether the interface itself is the obstacle. Starting from the G checkpoint (initialization 204, order 954), we train on a frozen Family-D target stream (data seed 643) under the \emph{restricted} target layout for exactly 5{,}000 updates, holding the LoRA initialization (the G checkpoint's trained LoRA), architecture, optimizer family, and budget fixed across three arms that differ only in the treatment of the communication interface (the packet reader and writer projections and the mouth):

\begin{itemize}
\item \textbf{Arm a (inherited, frozen):} reader/writer/mouth fixed at their G-trained state; LoRA trainable.
\item \textbf{Arm b (inherited, trainable):} the same inherited interface state, fully adaptable alongside the LoRA.
\item \textbf{Arm c (fresh interface):} reader/writer/mouth reinitialized; the same G-trained LoRA, all trainable.
\end{itemize}

Every arm is also evaluated with all packets cut, so that any success must be packet-mediated. The preregistered primary endpoints are final held-out depth-three accuracy and the normalized area under the learning curve (AUC).

\subsection{Results}

\begin{table}[t]
\centering
\caption{P3 factorial on the globally visible checkpoint (target stream d643, 5{,}000 updates). Chance $=0.0588$. All arms collapse to chance when communication is cut, so no arm escapes through direct evidence.}
\label{tab:p3}
\begin{tabular}{lccc}
\toprule
Arm & Final depth-3 & Normalized AUC & All-cut final \\
\midrule
a: inherited interface, frozen    & 0.0647 & 0.076 & 0.0588 \\
b: inherited interface, trainable & 0.1691 & 0.070 & 0.0588 \\
c: fresh interface, same LoRA     & \textbf{0.8574} & \textbf{0.435} & 0.0588 \\
\bottomrule
\end{tabular}
\end{table}

\begin{figure}[t]
\centering
\includegraphics[width=0.72\linewidth]{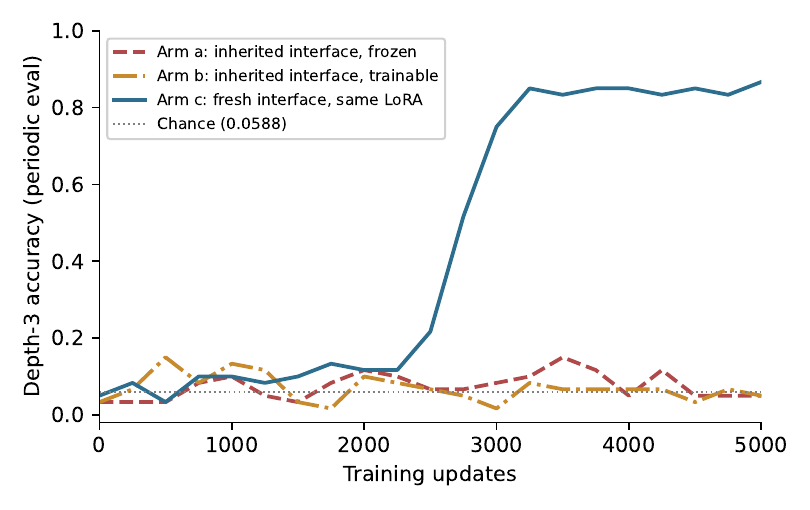}
\caption{P3 depth-three learning curves on the globally visible checkpoint (periodic evaluations every 250 updates; the endpoints in Table~\ref{tab:p3} use the larger full held-out bank). The frozen inherited interface (arm a) remains at chance, the fully trainable inherited interface (arm b) stays trapped, and the fresh interface with the same globally trained LoRA (arm c) breaks out sharply.}
\label{fig:p3curves}
\end{figure}

The preregistered fresh-interface advantage passes by very large margins: $\Delta_{\mathrm{final}}(c-a)=+0.7927$ and $\Delta_{\mathrm{AUC}}(c-a)=+0.359$; $\Delta_{\mathrm{final}}(c-b)=+0.6883$ and $\Delta_{\mathrm{AUC}}(c-b)=+0.365$ (Table~\ref{tab:p3}, Figure~\ref{fig:p3curves}).

In the sole high-performing globally visible checkpoint, the inherited communication interface produced severe negative transfer on Family D. With the same globally trained LoRA, target stream, restricted target layout, and training budget, reinitializing only the reader, writer, and mouth raised final depth-three accuracy from 0.065 with the frozen inherited interface and 0.169 with the fully adaptable inherited interface to 0.857. All three arms remained communication-dependent. Thus, within this matched factorial, the inherited global interface---not an inability of the LoRA to learn the target family---was the decisive source of failure.

The full-adaptation arm matters most. This was not merely a frozen-interface compatibility problem: even with the inherited reader, writer, and mouth fully trainable, the system remained trapped at 0.169 within 5{,}000 updates, while reinitialization yielded a sharply different optimization trajectory, consistent with basin dependence. We therefore describe the result as \emph{interface-level negative transfer}: the inherited interface initialization was the decisive bottleneck in the P3 factorial. The experiment identifies the treatment responsible for the b-versus-c contrast; it does not prove that no other limitation exists in the checkpoint, nor that every globally trained interface would behave similarly.

Conversely, the 0.857 endpoint---descriptively the strongest 5{,}000-update result in the campaign---shows that the globally trained LoRA retained strong capacity to acquire Family-D operator semantics once decoupled from its inherited interface. Whether global visibility systematically improves local-skill acquisition remains untested: the comparisons to restricted arms differ in checkpoint, initialization, and target stream, and P3 contains one global model.

\section{P4: second-stream replication of the inheritance comparison}\label{sec:p4}

The scout's interface-inheritance probe compared inherited (arm b) against fresh (arm c) interfaces for two restricted checkpoints on one target stream each, and found no preregistered large benefit. Because that null could have been stream-specific, P4 repeats the b/c comparison on a second independently frozen target stream per checkpoint, with the criterion unchanged: inherited must exceed fresh by $+0.10$ in both normalized AUC and final depth-three accuracy for both tested checkpoints.

\begin{table}[t]
\centering
\caption{P4: inherited-minus-fresh differences ($b-c$) across two restricted checkpoints and two independently frozen target streams each. The preregistered $+0.10$ criterion is never met.}
\label{tab:p4}
\begin{tabular}{llrr}
\toprule
Checkpoint & Stream & $\Delta$ final ($b-c$) & $\Delta$ AUC ($b-c$) \\
\midrule
\ckpt{m201\_o951} & d640 (original) & $-0.0206$ & $+0.0012$ \\
\ckpt{m201\_o951} & d644 (new)      & $-0.0417$ & $+0.0392$ \\
\ckpt{m203\_o953} & d641 (original) & $+0.0201$ & $+0.0421$ \\
\ckpt{m203\_o953} & d645 (new)      & $+0.0245$ & $+0.0825$ \\
\bottomrule
\end{tabular}
\end{table}

The preregistered large initialization-benefit criterion failed again under the second independently frozen data stream for each checkpoint (Table~\ref{tab:p4}). Across two restricted checkpoints and two streams per checkpoint, the inherited interface never exceeded the fresh-interface control by the required 0.10 margin in either normalized learning-curve area or final depth-three accuracy. The failure of the large-benefit criterion is therefore stream-replicated. This does not establish an exact null: AUC differences were consistently small and nonnegative, ranging from 0.001 to 0.083, while final differences were mixed, from $-0.042$ to $+0.025$. Small early-learning effects remain compatible with the data; what is excluded is the preregistered large advantage. We accordingly describe the outcome as \emph{the absence of a preregistered large initialization benefit, replicated across streams}, not as a replicated null. P4 serves as robustness evidence for the interpretation of P3, as a preregistered negative result, and as motivation for studying target families whose interfaces are more expensive to reinvent.

\section{Discussion}\label{sec:discussion}

\paragraph{Two canonicalities.} The campaign separates two properties that the parent paper's within-checkpoint audits could not distinguish. A packet interface is \emph{semantically canonical} within a checkpoint when packets from different episodes are interchangeable whenever they carry the same running value; the restricted interfaces have this property and the global interface does not. Two societies are \emph{coordinate-canonical} when raw packets exchange across checkpoints. P1 shows that semantic canonicality does not imply a shared coordinate convention: the restricted models learn portable semantic interfaces expressed in initialization-anchored coordinate dialects, not one universal raw vector language. Appendix~A adds a post hoc cross-regime case: a same-lineage twin pair is raw-incompatible at the first two interfaces but partially interoperable at the final interface, showing that coordinate compatibility can vary by relay stage and recipient context even when parameter initialization and underlying episode order are shared.

\paragraph{What links canonicality to reuse.} P3 ties \emph{semantic} canonicality---not raw cross-model coordinate identity---to transfer value. The combined experiments support a mechanistic link between semantic canonicality and interface reusability. Restricted-visibility training produced approximately value-indexed interfaces in all six audited checkpoints; in the two purposively selected checkpoints tested for adaptation, the reader/writer/mouth interface could remain frozen while the shared LoRA learned the new instruction family. The sole successful globally visible model in the parent cohort instead learned an episode-entangled interface that was neither transplantable across the new context nor reusable during adaptation; reinitializing that interface while retaining the same LoRA removed the failure. This is a strong within-programme dissociation, but because the global side contains one checkpoint, it does not establish a population-level law that canonical interfaces always transfer or that globally trained interfaces generally cause negative transfer. Canonicality and reusability dissociate together in the present mechanistic case study.

\paragraph{Consolidated statement.} A preregistered follow-up campaign separated semantic portability, coordinate alignment, and adaptation value. All six restricted-visibility-trained interfaces transported the Family-A value code perfectly into a new linguistic and operator context; in the two purposively selected checkpoints tested for adaptation, the interface could remain frozen while the shared LoRA learned the new family. Yet these interfaces did not form one raw cross-model language: exact interoperability occurred only for one same-initialization pair, another same-initialization pair was partially compatible, and no cross-initialization direction passed the frozen alignment ladder. Moreover, inherited restricted interfaces conferred no preregistered large advantage over fresh interfaces across two checkpoints and two target streams each. The sole successful globally visible interface in the parent cohort showed the opposite failure mode: it was episode-entangled, nonportable, and induced severe negative transfer, while reinitializing only that interface enabled 0.857 depth-three accuracy with the same globally trained LoRA. All six restricted checkpoints also parsed and published the sealed Family-D start value perfectly; their zero-shot failure was confined, within the tested decomposition, to interpretation and execution of the new operator instructions. Together, the results indicate that restricted visibility favors semantically reusable value carriers expressed in initialization-anchored coordinate dialects, whereas the observed globally trained interface acted as a harmful prior for new-family learning.

\paragraph{Limitations.} These conclusions remain bounded to a near-transfer 17-state setting. P3 uses one globally visible checkpoint, one independently frozen target stream, and one fixed 5{,}000-update adaptation budget. The initialization stratification rests on two initializations represented under two data orders each, one pair only partially compatible; it is a sample-bound functional observation, not a causal law that initialization determines the protocol. The alignment ladder bounds only the tested map families (raw, orthogonal, affine, width-64 nonlinear); richer alignments were not tested. P2's localization is bounded to a single sealed start-value formulation. P4 excludes only the preregistered large benefit, not small effects. The 17-state interface is cheap to reinvent, which plausibly caps any inheritance benefit; whether inheritance pays when reinvention is expensive is precisely the open question the negative and null results here motivate. Appendix~A is post hoc, success-conditioned, and examines the sole task-successful GT trajectory from a three-run tagged-global pilot; it does not estimate the frequency of cross-regime interoperability, establish semantic use of the ownership tags, or isolate visibility from tagging.

\paragraph{Outlook.} The natural next experiment treats inheritance benefit as a function of target reinvention cost and source diversity: source-pretrained interfaces evaluated against matched fresh-interface controls on targets with larger or structured state spaces, with positive, neutral, and negative interface transfer all classified prospectively---P3 shows the negative outcome is real and must be a first-class hypothesis rather than an implementation failure.

\section*{Reproducibility and artifacts}
All experiments ran on the parent study's released checkpoints, pinned genome revision, and pinned software stack, on machines admitted by bit-exact retrace of a golden training segment and matching fresh-world fingerprints. The P1 launch manifest---split, derangement, programme, and pairing seeds; mapper seeds; checkpoint SHA-256 hashes; and the hash of the frozen classification script---was sealed before any held-out score was computed, and the sealed manifest additionally records bank and map hashes. Evaluation outputs, sealed manifests, bank-construction and evaluation scripts, and the P2--P4 configurations and results accompany the parent study's evidence release (\url{https://github.com/tokenosopher/populus-evidence-partitioning}). The twin-dialect release includes RNG seeds 8300 and 8400, donor and recipient episode metadata, per-case predictions, and the complete self/cross transplant matrix. Model weights are hosted in the companion Hugging Face repository (\url{https://huggingface.co/tokenosopher/populus-evidence-partitioning-checkpoints}): all twenty parent-society final checkpoints, the tagged-global pilot checkpoint, the P3/P4 adapted-run finals, and the fitted P1 alignment maps (\texttt{p1\_maps.pt}).

\appendix

\section{Post hoc case study: twin dialects across training regimes}\label{app:twins}

P1 varied initialization and data order \emph{within} one training regime and found initialization-anchored coordinate dialects. This appendix reports a complementary, post hoc case study on the orthogonal axis: one pair of societies with byte-identical parameter initialization and the same ordered sequence of underlying task episodes, but trained under \emph{different information regimes}. Because the role-tagged model prepends ownership markers to its spans, the serialized token streams are not byte-identical. The study was run after the P1--P4 campaign, outside the sealed protocol, and concerns a single pair; we include it because it cleanly separates two properties that the campaign's within-regime audits could not---and because its instrument, raw cross-system transplantation with counterfactual scoring and recipient self-transplant normalization, is preregistered for reuse in the follow-up cohort.

\paragraph{Setup.}
The parent programme's exploratory role-marked pilot trained three globally visible societies whose evidence spans carried single-token ownership markers (``mine''/``other'' prepended to each span, per cell). Only GT~\ckpt{m200\_o950} attained high depth-three held-out accuracy; the other two tagged-global trajectories remained low at 0.079 and 0.119. We examine that sole task-successful GT trajectory because cross-regime packet semantics are only interpretable in a model that performs the downstream task.

GT~\ckpt{m200\_o950} shares byte-identical parameter initialization and the same ordered sequence of underlying task episodes with the audited restricted checkpoint \ckpt{m200\_o950}. The regimes differ jointly in evidence visibility and tagging, so the contrast is regime-level rather than a visibility-only manipulation. The restricted twin scores 0.8775/0.6713 at depths two/three; the tagged-global twin scores 0.1593/0.7914, a deep-only inversion. In a post hoc collision-stratified rescore, the tagged-global twin scores 0.7636 on the 13 map-novel depth-three programs while remaining poor at depth two (0.1225 on the map-novel stratum), so its depth-three result is not explained by exact lookup of a complete affine map seen in training.

No marker-permutation audit was performed, so this case study treats GT as a tagged training regime and does not infer that it used the words ``mine'' and ``other'' according to their intended ownership semantics. The transplant matrix was executed on a golden-retrace-verified host. The restricted checkpoint is the released parent final; GT was trained for 20{,}000 updates under the matched optimization and sampling protocol on the same task world and underlying episode order.

\paragraph{Method.}
We harvested natural packets from each twin on its own correctly answered depth-three episodes, covering all 17 values and both diagnostic phrasing rotations, with at least four packets per interface--value pair. Donor harvesting used RNG seed 8300 and recipient-episode selection used RNG seed 8400. Interfaces $k\in\{0,1,2\}$ denote the first, second, and third inter-cell packet boundaries. Substitution replaces the recipient's committed packet after the corresponding cell update, matching the parent audit convention.

For each recipient and interface, the same 120 episodes that the recipient answered correctly before intervention were used for both donor models; intact accuracy on these success-conditioned banks is therefore 1.000 by construction. Raw donor packets were injected with no fitted map under four interventions, yielding five reported scores: (i) a same-value donor packet; (ii) a counterfactual donor packet carrying $v'=(v{+}3)\bmod 17$, scored both against the suffix-implied counterfactual target (\emph{counterfactual following}) and against the original answer (\emph{shifted retention}); (iii) a random mismatched-value donor packet carrying $w\neq v$, scored for retention of the original answer; and (iv) zero-packet deletion. Chance is $1/17\approx0.0588$. Per-case predictions and donor metadata are released with the evidence artifacts.

An artifact-level audit of donor--recipient episode identity, keyed on the ordered program, start value, and phrasing rotation, found 11 of 1{,}440 same-value transplants (at most 3 per 120-episode row) in which the donor packet originated from the exact recipient episode. Excluding these cases changes no reported score by more than 0.007; in particular, the restricted twin's self-interchange remains 1.000 at all three interfaces ($n=118$--$119$ after exclusion). Donors sharing the recipient's ordered program under any start value numbered 0--4 per row.

\begin{table}[t]
\centering
\small
\caption{Twin-dialect $2\times2$ transplant matrix. Accuracy over $n=120$ recipient episodes per recipient--interface row, each naturally answered correctly before intervention (intact accuracy 1.000 by construction). Raw donor packets are substituted at interface $k$. ``Cf.'' scores the suffix-implied counterfactual target; shifted retention and mismatched-value score retention of the recipient episode's original answer. Chance $\approx0.0588$. P $=$ restricted twin and GT $=$ role-tagged global twin, both from the \ckpt{m200\_o950} lineage.}
\label{tab:twins}
\begin{tabular}{llccccc}
\toprule
Recipient & Donor & $k$ & Same value & Cf.\ following & Shifted retention & Mismatched value \\
\midrule
P  & P  & 0 & 1.000 & 1.000 & 0.000 & 0.000 \\
P  & P  & 1 & 1.000 & 1.000 & 0.000 & 0.000 \\
P  & P  & 2 & 1.000 & 1.000 & 0.000 & 0.058 \\
P  & GT & 0 & 0.067 & 0.100 & 0.058 & 0.033 \\
P  & GT & 1 & 0.050 & 0.092 & 0.017 & 0.058 \\
P  & GT & 2 & 0.567 & 0.650 & 0.008 & 0.058 \\
GT & GT & 0 & 0.133 & 0.067 & 0.092 & 0.092 \\
GT & GT & 1 & 0.183 & 0.183 & 0.075 & 0.075 \\
GT & GT & 2 & 0.492 & 0.425 & 0.042 & 0.033 \\
GT & P  & 0 & 0.042 & 0.050 & 0.042 & 0.067 \\
GT & P  & 1 & 0.075 & 0.058 & 0.083 & 0.067 \\
GT & P  & 2 & 0.308 & 0.275 & 0.058 & 0.042 \\
\bottomrule
\end{tabular}

\smallskip
\raggedright\footnotesize Deletion controls (zero packet, per recipient and interface): P 0.050/0.050/0.067; GT 0.083/0.033/0.050.
\end{table}

\paragraph{Findings.} Four observations survive adjudication (Table~\ref{tab:twins}). \emph{First}, the restricted twin's packet was an exact causal sufficient statistic for the running value with respect to the downstream answer under the tested suffixes, at all three interfaces on the success-conditioned audit bank: natural same-value substitutions preserved behavior, counterfactual-value substitutions redirected every prediction to the mathematically implied answer, and deletion or random value mismatch reduced performance to chance or near chance. (Interface 0 is structurally simpler because the source cell's outgoing state is largely determined by the start value; interfaces 1 and 2 are the stronger evidence.) \emph{Second}, the tagged-global twin's packet semantics are strongly episode-dependent: its own natural packets are not reliably interchangeable across episodes, even at matched running values (0.133/0.183/0.492), although deletion still collapses performance. The channel is causally necessary without being semantically canonical---the same qualitative dissociation the parent study observed in its untagged global outlier, now in the sole task-successful role-tagged global trajectory from a three-run exploratory pilot. Its interface-2 packets are partially value-aligned but context-welded, not uninterpretable. \emph{Third}, the twins' early packet interfaces were functionally non-interoperable raw codes in this pair, while the final inter-cell interface retained partial bidirectional interoperability. \emph{Fourth}, the large raw directional asymmetry is consistent with a predominantly recipient-side explanation. Descriptively normalizing each cross-regime counterfactual score by the same recipient's self-interchange \emph{baseline} on independently sampled donor packets (with exact donor--recipient overlap audited separately) yields nearly symmetric late-interface compatibility: $0.650/1.000=0.650$ versus $0.275/0.425\approx0.647$. The tagged-global recipient reads predominantly cross-episode packets poorly, including its own. This ratio analysis is not a formal decomposition of donor and recipient effects; donor-bank quality and donor--recipient interactions remain possible, but the large unnormalized directionality is primarily associated with the recipients' very different self-interchange baselines.

\paragraph{Interpretation and scope.}
The dissociation is consistent with a well-established prediction from communication research: predictable or shared context permits less autonomous, more context-dependent signals, whereas receivers lacking that context place greater pressure on a signal to carry self-sufficient meaning \citep{winters2018contextual,piantadosi2012ambiguity,glowka2024context,gualdoni2024bridging}. At the level of coding intuition, decoder side information can reduce what the transmitted code itself must carry \citep{slepian1973noiseless,wyner1976rate}; the present learned neural channel is not a direct test of those coding theorems.

In this one neural twin pair, the restricted twin's packets were value-canonical and interchangeable across the audited correctly answered episodes, whereas the tagged-global twin's packets remained strongly context-dependent. The late-interface compatibility should not be attributed to initialization alone: the pair shares initialization, underlying episode order, architecture, terminal task, and output decoding, and the present design does not separate these common pressures. The result establishes partial raw functional compatibility at the final interface; it does not establish a single shared packet language or measure geometric alignment between the two spaces.

\paragraph{Novelty boundary.}
Prior work studies incompatible emergent conventions and cross-play failure \citep{hu2020otherplay,bullard2021qed,wolski2026crossplay}, cross-model latent alignment and stitching \citep{lenc2014equivariance,bansal2021stitching,moschella2022relative,peng2026statebridge}, causal activation interchange within and across models \citep{geiger2021causal,grant2025mas,cardenas2026causal}, raw latent non-transferability across seeds \citep{okatan2025keys}, recipient-side determinants of transferred-state utility \citep{li2026memory}, translation between emergent protocols \citep{levy2025translation}, and context-dependent signaling \citep{winters2018contextual,glowka2024context,gualdoni2024bridging}.

To our knowledge, we found no prior study combining same-lineage cross-regime training---defined here as shared parameter initialization and underlying episode order---with raw cross-system transplantation of an emergent inter-agent message, exact semantic counterfactual scoring, and recipient-normalized self-interchange analysis. Nor did we find a prior report of this exact interface-resolved pattern for raw emergent inter-agent messages under same-lineage, cross-regime training. Because this analysis concerns one pair, one synthetic state space, success-conditioned episodes, and a post hoc-selected task-successful GT trajectory, it is an exploratory mechanistic case study rather than a population claim. The preregistered follow-up cohort will apply the same matrix audit to every matched twin pair.

\bibliographystyle{plainnat}
\bibliography{references}

\end{document}